\def\year{2020}\relax

\documentclass[letterpaper]{article} 
\usepackage{cite}
\usepackage{aaai20}  
\usepackage{times}  
\usepackage{helvet} 
\usepackage{courier}  
\usepackage[hyphens]{url}  
\usepackage{graphicx} 
\usepackage{graphicx}  
\usepackage{graphics}    
\usepackage{times}       
\usepackage{mathrsfs}

\usepackage{amsmath}     
\usepackage{amssymb}     
\usepackage[noend]{algorithmic}
\usepackage{algorithm} 

\usepackage[shortcuts,acronym]{glossaries}
 
\usepackage{amsthm}
\usepackage{thmtools}

\usepackage[T1]{fontenc}

\usepackage{graphicx}
\usepackage{subcaption}

\usepackage{tikz-cd}
\usetikzlibrary{shapes}
\usetikzlibrary{arrows.meta}
\usetikzlibrary{arrows}
\usetikzlibrary{positioning}
\usetikzlibrary{fadings}
\tikzset{
  shift left/.style ={commutative diagrams/shift left={#1}},
  shift right/.style={commutative diagrams/shift right={#1}}
}

\definecolor{colorborder}{RGB}{255,203,169}
\definecolor{colorinside}{RGB}{255,245,231}

\usepackage{color}
\definecolor{orange}{rgb}{1.0, 0.4, 0.0}

\newtheoremstyle{mainstyle}
  {\topsep} 
  {\topsep} 
  {\itshape} 
  {} 
  {\bfseries \itshape} 
  {.} 
  {.5em} 
  {} 
\newtheoremstyle{proofstyle}
  {\topsep} 
  {\topsep} 
  {\itshape} 
  {} 
  {\itshape} 
  {:} 
  {.5em} 
  {} 

\theoremstyle{mainstyle}
\newtheorem{myassumption}{Assumption} \def\assref#1{Assumption~\ref{#1}}
 
\newtheorem{mydefinition}{Definition} \def\defref#1{Definition~\ref{#1}}
\newtheorem{mylemma}{Lemma} \def\lemref#1{Lemma~\ref{#1}}
\newtheorem{myremark}{Remark} 
\newtheorem{myproposition}{Proposition} \def\propref#1{Proposition~\ref{#1}}
\newtheorem{mycorollary}{Corollary} 

\theoremstyle{proofstyle}
\newtheorem{myproof}{Proof} 
\newcommand*{\qedproof}{\hfill\ensuremath{\square}}%

\def\algref#1{Algorithm~\ref{#1}}
\def\secref#1{Section~\ref{#1}}
\def\figref#1{Fig.~\ref{#1}}

\def\eqref#1{Eq.~(\ref{#1})}

\newcommand\etal{\emph{et al. }}

\newcommand\ADGgraph{$\mathcal{G}_\text{ADG}$ }

\title{\LARGE \bf A Feedback Scheme to Reorder a Multi-Agent Execution Schedule by Persistently Optimizing a Switchable Action Dependency Graph}

\author{Alexander Berndt%
\thanks{Alexander Berndt and Tam\'as Keviczky are with the Delft Center for Systems and Control (DCSC),
TU Delft, 2628 CN Delft, The Netherlands
{\tt\small berndtae@gmail.com, T.Keviczky@tudelft.nl}.}
\And Niels van Duijkeren%
\thanks{Niels van Duijkeren and Luigi Palmieri are with Robert Bosch GmbH, Corporate Research,
Renningen, 71272, Germany
{\tt\small \{Niels.vanDuijkeren, Luigi.Palmieri\}@de.bosch.com}.}
\And Luigi Palmieri\footnotemark[2]
\And Tam\'as Keviczky\footnotemark[1]{\rm}
}

\providecommand{\keywords}[1]{\textbf{\textit{Index terms---}}\textit{#1}}

\newacronym{ADG}{ADG}{Action Dependency Graph}
\newacronym{SADG}{SADG}{Switchable Action Dependency Graph}
\newacronym{AGV}{AGV}{Automated Guided Vehicle}
\newacronym{AGVs}{AGVs}{Automated Guided Vehicles}
\newacronym{CBC}{CBC}{Coin-Or Branch-and-Cut}
\newacronym{CBS}{CBS}{Conflict-Based Search}
\newacronym{CCBS}{CCBS}{Continuous Conflict-Based Search}
\newacronym{DMAPF}{D-MAPF}{Dynamic Multi-Agent Path Finding}
\newacronym{ECBS}{ECBS}{Enhanced Conflict-Based Search}
\newacronym{MAPF}{MAPF}{Multi-Agent Path Finding}
\newacronym{MAPD}{MAPD}{Multi-Agent Pickup and Delivery}
\newacronym{MILP}{MILP}{Mixed-Integer Linear Program}
\newacronym{MILPs}{MILPs}{Mixed-Integer Linear Programs}
\newacronym{SIPP}{SIPP}{Safe Interval Path Planning}

\begin{document}
\maketitle
\thispagestyle{empty}
\pagestyle{empty}
\begin{abstract}
  In this paper we consider multiple \ac{AGVs} navigating a common workspace to fulfill various intralogistics tasks, 
    typically formulated as the \ac{MAPF} problem.
  To keep plan execution deadlock-free, 
    one approach is to construct an \ac{ADG} 
    which encodes the ordering of \acs{AGVs} as they proceed along their routes.
  Using this method, 
    delayed \acs{AGVs} occasionally require others to wait for them at intersections,
    thereby affecting the plan execution efficiency.
  If the workspace is shared by dynamic obstacles such as humans or third party robots, 
    \acs{AGVs} can experience large delays.
  A common mitigation approach is to re-solve the \acs{MAPF} using the current, 
    delayed \acs{AGV} positions.
  However, solving the \acs{MAPF} is time-consuming, 
    making this approach inefficient,
    especially for large \acs{AGV} teams. 
  In this work, we present an online method to repeatedly modify a given acyclic \acs{ADG} 
    to minimize the cumulative \acs{AGV} route completion times.
  Our approach persistently maintains an acyclic \acs{ADG}, 
    necessary for deadlock-free plan execution.
  We evaluate the approach by considering simulations with random disturbances on the execution and  
    show faster route completion times compared to 
    the baseline \acs{ADG}-based execution management approach.
\end{abstract}

\keywords{
Robust Plan Execution, 
Scheduling and Coordination, 
Mixed Integer Programming, 
Multi-Agent Path Finding, 
Factory Automation}

\section{Introduction}
\label{sec:intro}  
 
Multiple \acf{AGVs} have shown to be capable of efficiently performing intra-logistics 
  tasks such as moving inventory in 
  distribution centers \cite{wurmanCoordinatingHundredsCooperative2008}.
The coordination of \ac{AGVs} in shared environments is typically formulated as the \acf{MAPF} problem, 
  which has been shown to be NP-Hard \cite{yuMultiagentPathPlanning2013}.
The problem is to find trajectories for each \acs{AGV} along a roadmap such that 
  each \acs{AGV} reaches its goal without colliding with the other \ac{AGVs}, 
  while minimizing the makespan. 
The \acs{MAPF} problem typically considers an abstraction of the workspace to a graph 
  where vertices represent spatial locations 
  and edges pathways connecting two locations.
  
\begin{figure}
	\centering
	\includegraphics[width=0.87\linewidth]{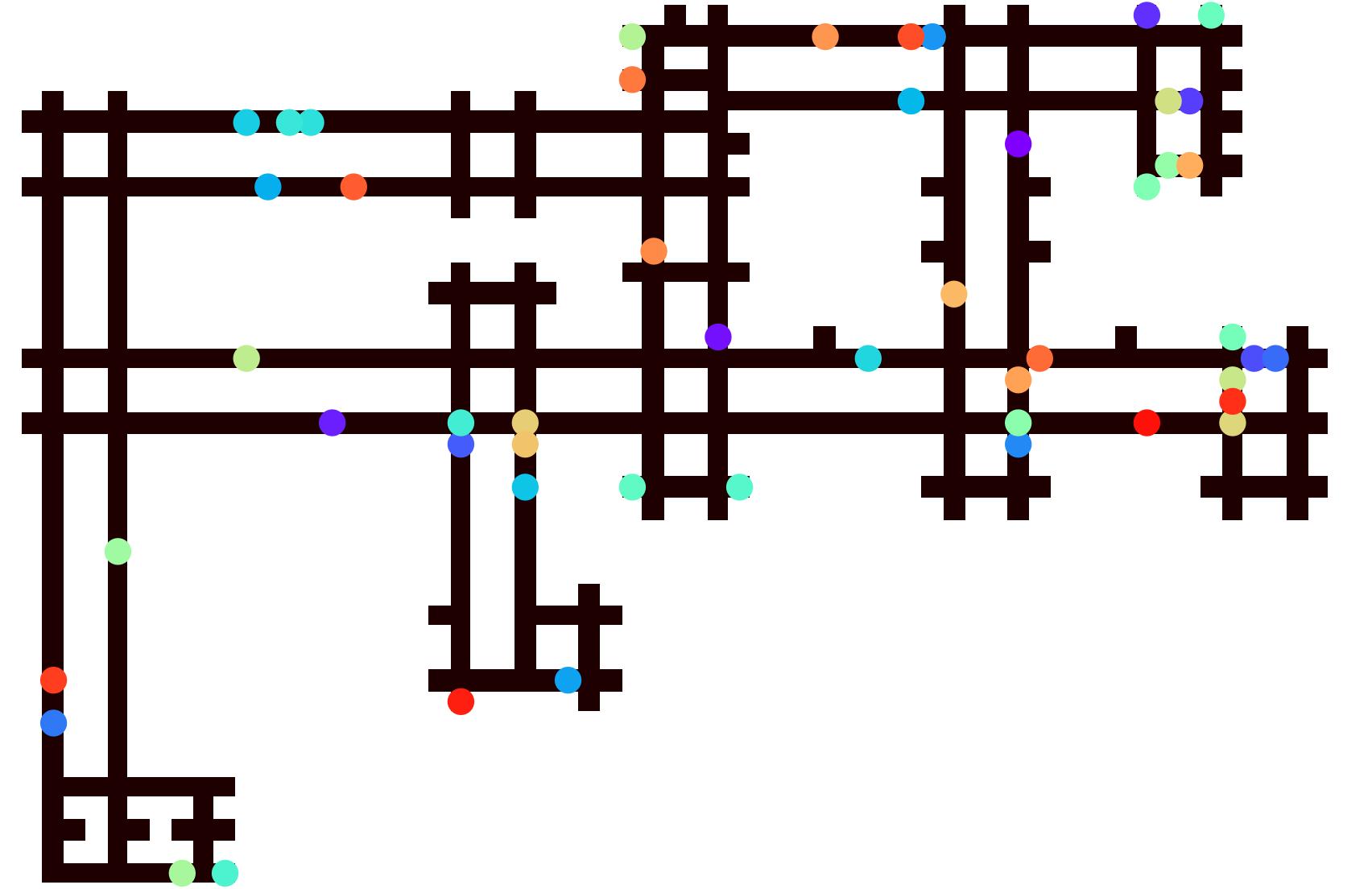}
	\caption{A roadmap occupied by 50 \ac{AGVs} 
		(represented by colored dots). 
		\acs{AGVs} must efficiently navigate from a start to a goal position 
		while avoiding collisions with one another, 
		despite being subjected to delays.}
	\label{fig:agvs_on_map}
\end{figure}

Recently, solving the \acs{MAPF} problem has garnered widespread 
  attention 
  \cite{sternMultiAgentPathfindingDefinitions2019,felnerSearchBasedOptimalSolvers2017}. 
This is mostly due to the abundance of application domains, such as intralogistics, 
  airport taxi scheduling \cite{morris2016planning} and computer games \cite{mOntanon2013survey}. 
Solutions to the \acs{MAPF} problem include \ac{CBS} \cite{sharonConflictbasedSearchOptimal2015}, 
  Prioritized Planning using \ac{SIPP} \cite{DBLP:journals/corr/YakovlevA17}, 
  declarative optimization approaches using answer set programming \cite{bogatarkanDeclarativeMethodDynamic2019}, 
  heuristic-guided coordination \cite{pecora2018loosely} 
  and graph-flow optimization approaches \cite{yuPlanningoptimalpaths2013}.

Algorithms such as \acs{CBS} have been improved 
  by exploiting properties such as geometric symmetry \cite{liSymmetryBreakingConstraintsGridBased2019},
  using purpose-built heuristics \cite{felnerAddingHeuristicsConflictBased2018}, 
  or adopting a \ac{MILP} formulation 
  where a branch-cut-and-price solver is used to 
  yield significantly faster solution times \cite{lamBranchandCutandPriceMultiAgentPathfinding2019}.
%

Similarly, the development of bounded sub-optimal solvers 
  such as \ac{ECBS} \cite{barerSuboptimalVariantsConflictBased2014} 
  have further improved planning performance for higher dimensional state spaces. 
\ac{CCBS} can be used to determine \acs{MAPF} plans for more realistic roadmap 
  layouts \cite{andreychukMultiAgentPathfindingContinuous2019}. 
As opposed to \acs{CBS}, 
  \acs{CCBS} considers a weighted graph and continuous 
  time intervals to describe collision avoidance constraints, 
  albeit with increased solution times.
%

The abstraction of the \acs{MAPF} to a graph search problem means
  that executing the \acs{MAPF} plans requires monitoring of the assumptions made 
  during the planning stage to ensure and maintain their validity. 
This is because irregularities such as 
  vehicle dynamics and unpredictable delays influence plan execution.
$k$R-\acs{MAPF} addresses this by permitting delays 
  up to a duration of $k$ time-steps \cite{atzmonRobustMultiAgentPath2020}.
Stochastic \acs{AGV} delay distributions are considered in \cite{maMultiAgentPathFinding2017}, 
  where the \acs{MAPF} is solved by minimizing the expected overall delay.
These robust \acs{MAPF} formulations and solutions 
  inevitably result in more conservative plans compared to their nominal counterparts. 

An \acf{ADG} encodes the ordering between \acs{AGVs} 
  as well as their kinematic constraints in a post-processing step 
  after solving the \acs{MAPF} \cite{hoenigMultiAgentPathFinding2016}.
Combined with an execution management approach, this allows \acs{AGVs} 
  to execute \acs{MAPF} plans successfully despite 
  kinematic constraints and unforeseen delays.
This work was extended to allow for
  persistent re-planning \cite{hoenigPersistentRobustExecution2018}.

The aforementioned plan execution solutions in 
  \cite{hoenigPersistentRobustExecution2018,atzmonRobustMultiAgentPath2020,maMultiAgentPathFinding2017}   
  address the effects of delays by 
  ensuring synchronous behavior among \acs{AGVs} while 
  maintaining the originally planned schedule's ordering. 
The result is that plan execution is unnecessarily inefficient 
  when a single \acs{AGV} is largely delayed and others are on schedule, 
  since \acs{AGVs} need to wait for the delayed \acs{AGV} 
  before continuing their plans.
We observe that to efficiently mitigate the effects of large delays 
  the plans should be adjusted continuously in an online fashion,
  where the main challenges are to maintain the original plan's
  deadlock- and collision-free guarantees.
%


In this paper, 
  we present such an online approach capable of 
  reordering \acs{AGVs} based on a \ac{MAPF} solution, 
  allowing for efficient \ac{MAPF} plan execution 
  despite \acs{AGVs} being subjected to large delays.
This approach is fundamentally different from the 
  aforementioned approaches 
  \cite{hoenigMultiAgentPathFinding2016,maMultiAgentPathFinding2017,atzmonRobustMultiAgentPath2020,hoenigPersistentRobustExecution2018} 
  in that delays can be accounted for as they occur, 
  instead of anticipating them \textit{a priori}.
The feedback nature of our approach additionally means 
  solving the initial \acs{MAPF} can be done assuming nominal plan execution, 
  as opposed to solving a robust formulation 
  which necessarily results in plans of longer length 
  due to the increased conservativeness.
  
Our contributions include 
  an optimization formulation based on a novel \acf{SADG}
  to re-order \acs{AGV} dependencies.
Monte-Carlo simulation results show lower cumulative route completion times 
  with real-time applicable optimization times while guaranteeing
  collision- and deadlock-free plan execution.


Working towards our proposed solution, 
  we formally define the \acs{MAPF} problem and the concept of an \acs{ADG}
  in \secref{sec:preliminaries}.
Based on a modified version of this \acs{ADG}, 
  we introduce the concept of a reverse agent dependency in \secref{sec:switching_dependencies}.
This will allow an alternative ordering of \acl{AGVs}, 
  while maintaining a collision-free schedule.
In \secref{sec:opt_formulation}, 
  we formulate the choice of selecting between forward or reverse \acs{ADG} dependencies as 
  a mixed-integer linear programming problem.
The optimization problem formulation guarantees that the resulting \acs{ADG} 
  allows plan execution to be both collision- and deadlock-free, 
  while minimizing the predicted plan completion time.
Finally, we compare this approach to the 
  baseline \acs{ADG} method in \secref{sec:evaluation}.




%

\section{Preliminaries}
\label{sec:preliminaries}

Let us now introduce the fundamental concepts on which our approach is based,
  facilitated by the example shown in \figref{fig:example_illustration}.
Consider the representation of a workspace 
  by a roadmap $\mathcal{G} = (\mathcal{V},\mathcal{E})$,
  where $\mathcal{V}$ is a set of vertices and $\mathcal{E}$ a set of edges,
  e.g., as in \figref{fig:example_graph}.

\begin{mydefinition}[\acs{MAPF} Solution]
  The roadmap $\mathcal{G} = (\mathcal{V},\mathcal{E})$ 
    is occupied by a set of $N$ \acs{AGVs} 
    where the $i^\text{th}$ \acs{AGV} 
    has start $s_i \in \mathcal{V}$ 
    and goal $g_i \in \mathcal{V}$,
    such that $s_i \neq s_j$ and $g_i \neq g_j \; \forall \; i,j \in \{1,\ldots,N\}, \; i \neq j $.
  A \acs{MAPF} solution $\mathcal{P} = \{ \mathcal{P}_1,\dots,\mathcal{P}_N \}$ 
    is a set of $N$ plans, 
    each defined by a sequence $\mathcal{P} = \{ p^1, \dots, p^{N_i} \}$ of tuples $p = (l,t)$,
    with a location $l \in \mathcal{V}$ and a time $t \in [0, \infty)$.
  The MAPF solution is such that, 
    if every \acs{AGV} perfectly follows its plan, 
    then all \acs{AGVs} will reach 
    their respective goals in finite time without collision.     	 
  \label{def:MAPF_solution}
\end{mydefinition}

For a plan tuple $p = (l,t)$,
  let us define the operators $l = loc(p)$ and $t = \hat{t}(p)$ 
  which return the location $l \in \mathcal{V}$ 
  and \textit{planned} time 
  of plan tuple $p$ respectively. 
Let $S(l) \mapsto S \subset \mathbb{R}^2$ be an operator 
  which maps a location $l$ (obtained from $l = loc(p)$)
  to a spatial region in the physical workspace in $\mathbb{R}^2$.
Let $S_\text{AGV} \subset \mathbb{R}^2$ 
  refer to the physical area occupied by an \acs{AGV}.

In \figref{fig:example_graph}, $AGV_1$ and $AGV_2$ 
  have start and goal $s_1 = A$, 
  $g_1 = H$ and $s_2 = E$, 
  $g_2 = D$, respectively.
For this example, using \acs{CCBS} \cite{andreychukMultiAgentPathfindingContinuous2019} 
  yields $\mathcal{P} = \{\mathcal{P}_1,\mathcal{P}_2\}$ as
\begin{itemize}
  \item[] $\mathcal{P}_1 = \{ (A,0), (B,1.0), (C,2.2), (G,3.1), (H,3.9) \}$,
  \item[] $\mathcal{P}_2 = \{ (E,0), (F,1.1), (G,3.9), (C,4.8), (D,5.9) \}$.
\end{itemize}
Note the implicit ordering in $\mathcal{P}$, 
  stating $AGV_1$ traverses $C-G$ before $AGV_2$. 

\subsection{Modified \acl{ADG}}

Based on a \acs{MAPF} solution $\mathcal{P}$, 
  we can construct a modified version of the original \acf{ADG}, 
  formally defined in \defref{def:action_dependency_graph}.
This modified \acs{ADG} encodes the sequencing of \acs{AGV} 
  movements to ensure the plans are executed 
  as originally planned despite delays.
\begin{mydefinition}[\acl{ADG}]
	An \acs{ADG} is a directed graph 
		$\mathcal{G}_\text{ADG} = (\mathcal{V}_\text{ADG}, \mathcal{E}_\text{ADG})$ 
		where the vertices represent events 
		of an \acs{AGV} traversing 
		a roadmap $\mathcal{G}$.
	A vertex $v_i^k = (\{p_1,\dots,p_q\}, \emph{{status}}) \in \mathcal{V}_\text{ADG}$ 
    denotes the $k^\text{th}$ event of the $i^\text{th}$ \acs{AGV} 
    moving from $loc(p_1)$,
		via intermediate locations, 
		to $loc(p_q)$, 
		where $q \geq 2$ denotes the number of consecutive plan tuples 
		encoded within $v_i^k$.
		$\emph{{status}} \in \{\emph{{staged}}, \emph{{in-progress}}, \emph{{completed}} \}$.
  The edge $(v_i^k, v_{j}^{l}) \in \mathcal{E}_\text{ADG}$,
    from here on referred to as a dependency,  
    states that $v_{j}^{l}$ cannot be \emph{{in-progress}} 
    or \emph{{completed}} until $v_{i}^{k} = \emph{{completed}}$.
  An edge $(v_i^k, v_{j}^{l}) \in \mathcal{E}_\text{ADG}$
    is classified as \emph{Type 1} if $i = j$ and \emph{Type 2} if $i \neq j$.
    \label{def:action_dependency_graph}
\end{mydefinition}

Initially, 
  the \textit{status} of $v_i^k$ are $\textit{staged} \; \forall \; i,k$.
Let us introduce $plan(v_i^k)$ which returns
  the sequence of plan tuples $\{p_1,\dots,p_q\}$ for $v_i^k \in \mathcal{V}_\text{ADG}$. 
Let the operators $s(v_i^k)$ and $g(v_i^k)$ 
  return the start and goal vertices $loc(p_1)$ and $loc(p_q)$ 
  of vertex $v_i^k$ respectively and
  $\oplus$ denote the Minkowski sum.
We also differentiate between 
  \textit{planned} and \textit{actual} 
  \acs{ADG} vertex completion times.
Let $\hat{t}_s(v_i^k)$ and $\hat{t}_g(v_i^k)$ 
  denote the \textit{planned} time 
  that event $v_i^k \in \mathcal{V}_\text{ADG}$ 
  starts (\emph{status} changes from \emph{staged} to \emph{in-progress})
  and is completed (\emph{status} changes from \emph{in-progress} to \emph{completed}),
  respectively.  
An \acs{ADG} can be constructed from a plan 
  $\mathcal{P}$ using \algref{alg:adg_construction}.
\acs{AGVs} can execute their plans as originally 
  described by the \acs{MAPF} solution $\mathcal{P}$ 
  by adhering to the \acs{ADG}, defined next in \defref{def:adg_execution}.

\begin{mydefinition}[Executing \acs{ADG} based plans]
  \acs{AGVs} adhere to the \acs{ADG} if each \acs{AGV}
  only starts executing an \acs{ADG} event $v_i^k$ (status of $v_i^k$ changes from 
  \emph{staged} to \emph{in-progress}) if 
  all dependencies pointing to $v_i^k$ have \emph{status} = \emph{completed} 
  for all $v_i^k \in \mathcal{V}_\text{ADG}$.
  \label{def:adg_execution}
\end{mydefinition}

\begin{figure}
  \centering
  \begin{subfigure}{0.48\textwidth}
    \centering
    \includegraphics[width=0.7\textwidth]{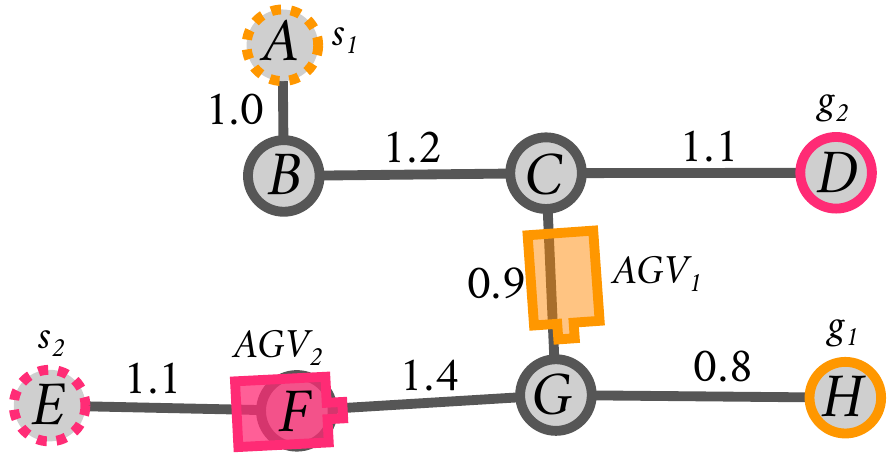}
    \caption{
      A roadmap graph occupied by two \acs{AGVs} with start $s_i$ and goal $g_i$ for $i = \{1,2\}$.
      The start and goal vertices are highlighted 
      with dotted and solid colored circle outlines respectively.
      The edge weights indicate the expected traversal times.
    }
    \label{fig:example_graph}
  \end{subfigure}
  \begin{subfigure}{0.48\textwidth}
    \centering
    \includegraphics[width=\textwidth]{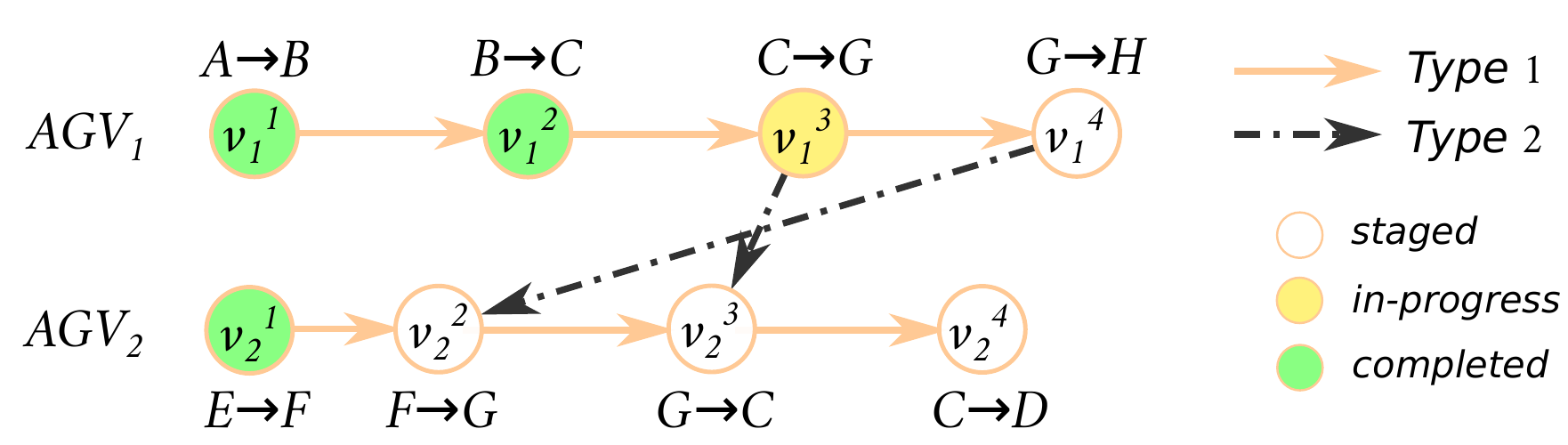}
    \caption{
      Illustration of the \acs{ADG} where each vertex status is color coded. It reflects the momentary progress of the \acs{AGVs} in \figref{fig:example_graph}.
    }
    \label{fig:example_ADG}
  \end{subfigure}
  \caption{Illustrative \acs{MAPF} problem example alongside the constructed \acl{ADG}}
  \label{fig:example_illustration}
\end{figure}

\figref{fig:example_ADG} shows 
  an example of \acs{AGVs} adhering to the \acs{ADG}.
Observe how $\acs{AGV}_2$ cannot start $v_2^2$ 
  before $v_1^4$ has been completed by $\acs{AGV}_1$, 
  as dictated by \defref{def:adg_execution}.

Next, we introduce \assref{ass:acyclic_ADG} 
  which we require to maintain deadlock-free
  behavior between \acs{AGVs} when executing an \acs{ADG} based plan
  as described in \defref{def:adg_execution}.
\begin{myassumption}[Acyclic \acs{ADG}]
  The \acs{ADG} constructed by \algref{alg:adg_construction} 
  using $\mathcal{P}$ as defined in \defref{def:MAPF_solution} 
  is acyclic.
	\label{ass:acyclic_ADG}
\end{myassumption}
\begin{myremark}
  \assref{ass:acyclic_ADG} can in practice always be satisfied
    in case the roadmap vertices outnumber the \acs{AGV} fleet size,
      i.e. $|\mathcal{V}| > N$ (as is typically the case in warehouse robotics). 
    Simple modificationa to existing \acs{MAPF} 
      solvers (e.g. an extra edge constraint in \acs{CBS}) is sufficient to obtain 
      \acs{ADG}s that satisfy \assref{ass:acyclic_ADG} \cite{hoenigPersistentRobustExecution2018}.
  \label{rem:acyclic_ADG_assumption}
\end{myremark}

\begin{algorithm}[]
  \caption{Modified \acs{ADG} construction based on \cite{hoenigPersistentRobustExecution2018}}
  \begin{algorithmic}[1]
    \renewcommand{\algorithmicrequire}{\textbf{Input:}}
    \renewcommand{\algorithmicensure}{\textbf{Result:}}
    \REQUIRE \acs{MAPF} solution $\mathcal{P} = \{ \mathcal{P}_1,\dots,\mathcal{P}_N \}$
    \ENSURE  $\mathcal{G}_\text{ADG}$ \\
    \vspace{2mm}
    // Add \acs{ADG} vertices and \textit{Type 1} dependencies
    \FOR {$i = 1$ to $N$}   \label{alg:adg_type_1_begin}
      \STATE $p \leftarrow p_i^1$
      \STATE $v \leftarrow (\{p\}, staged)$ 
      \STATE $v_\text{prev} \leftarrow None$
      \FOR {$k = 2$ to $N_i$}
        \STATE Append $p_i^k$ to $plan(v)$
        \IF {$S(loc(p)) \oplus S_\text{AGV} \cap S(loc(p_i^k)) \oplus S_\text{AGV} = \emptyset$}
        \label{alg:adg_construction_spatial_exclusivity}
        \STATE Add $v$ to $\mathcal{V}_\text{ADG}$
        \IF {$v_\text{prev}$ not \textit{None}}
          \STATE Add edge $(v_\text{prev}, v)$ to $\mathcal{E}_\text{ADG}$
        \ENDIF
        \STATE $v_\text{prev} \leftarrow v$
        \STATE $p \leftarrow p_i^k$
        \STATE $v \leftarrow (\{p\}, staged)$  \label{alg:adg_type_1_end}
      \ENDIF
      \ENDFOR
    \ENDFOR
    \vspace{2mm}
    // Add \textit{Type 2} dependencies
    \FOR {$i = 1$ to $N$} 
      \label{alg:adg_create_type_2}
      \FOR {$k = 1$ to $N_i$}
        \FOR {$j = 1$ to $N$}
          \IF {$i \neq j$}
            \FOR {$l = 1$ to $N_{j}$} \label{alg:adg_second_for_loop}
              \IF {$s(v_i^k) = g(v_j^l)$ \AND $\hat{t}_g(v_i^k) \leq \hat{t}_g(v_j^l)$}
              	\label{alg:type_2_dependency_condition}
                \STATE Add edge $(v_i^k, v_j^l)$ to $\mathcal{E}_\text{ADG}$
                \label{alg:adg_add_edge}
              \ENDIF
            \ENDFOR
          \ENDIF
        \ENDFOR
      \ENDFOR
    \ENDFOR
    \vspace{2mm}
    \RETURN $\mathcal{G}_\text{ADG}$
  \end{algorithmic}
  \label{alg:adg_construction}
\end{algorithm}



Unlike the originally proposed \acs{ADG} algorithm,
  \algref{alg:adg_construction} ensures that non-spatially-exlusive subsequent plan tuples
  are contained within a single \acs{ADG} vertex, 
  cf. line~\ref{alg:adg_construction_spatial_exclusivity} of the algorithm. 
This property will prove to be useful 
  with the introduction of reverse dependencies 
  in \secref{sec:reverse_dependencies}.
Despite these modifications, 
  \algref{alg:adg_construction} maintains the original algorithm's
  time complexity of $\mathcal{O}(N^2 \bar{n}^2)$ 
  where $\bar{n} = \max_i N_i$.

Due to delays, the \textit{planned} 
  and \textit{actual} \acs{ADG} vertex times 
  may differ.
Much like the previously introduced 
  \textit{planned} event start and completion times 
  $\hat{t}_s(v_i^k)$ and $\hat{t}_g(v_i^k)$, 
  we also introduce
  $t_s(v_i^k)$ and $t_g(v_i^k)$ 
  which denote the \textit{actual} start 
  and completion times 
  of event $v_i^k \in \mathcal{V}_\text{ADG}$
  respectively.
Note that if the \acs{MAPF} solution
  is executed nominally, i.e. \acs{AGVs} 
  experience no delays, then
  $t_s(v_i^k) = \hat{t}_s(v_i^k)$ and
  $t_g(v_i^k) = \hat{t}_g(v_i^k)$ for all
  $v_i^k \in \mathcal{V}_\text{ADG}$.
Let us introduce an important property 
  of an \acs{ADG}-managed plan-execution scheme, 
  \propref{prop:collision_deadlock_free_plan_exec},
  concerning guarantees of successful plan execution.

\begin{myproposition}[Collision- and deadlock-free \acs{ADG} plan execution]
  Consider an \acs{ADG}, $\mathcal{G}_\text{ADG}$, 
    constructed from a \acs{MAPF} solution as defined in \defref{def:MAPF_solution}
    using \algref{alg:adg_construction},
    satisfying \assref{ass:acyclic_ADG}.
  If the \acs{AGV} plan execution 
    adheres to the dependencies in $\mathcal{G}_\text{ADG}$, 
  then, assuming the \acs{AGVs} are subjected to 
  a finite number of delays 
  of finite duration, 
  the plan execution will be collision-free and completed in finite time.
  \label{prop:collision_deadlock_free_plan_exec}
\end{myproposition}
\begin{myproof}
	Proof by induction.
	Consider that $AGV_i$ and $AGV_j$ 
	traverse a 
	common vertex $\bar{p} \in \mathcal{G}$ 
	along their plans
	$\mathcal{P}_i$ and $\mathcal{P}_j$, 
	for any $i,j \in \{ 1,\dots,N \}, i \neq j$.
	By lines \ref{alg:adg_type_1_begin}-\ref{alg:adg_type_1_end} of 
	\algref{alg:adg_construction}, 
	this implies
	$g(v_i^k) = s(v_j^l) = \bar{p}$ 
	for some
	$v_i^k, v_j^l \in \mathcal{V}_\text{ADG}$. 
	By lines 
	\ref{alg:adg_create_type_2}-\ref{alg:adg_add_edge} 
	of \algref{alg:adg_construction},
	common vertices of 
	$\mathcal{P}_i$ and $\mathcal{P}_j$ 
	in $\mathcal{G}$ 
	will result in a Type 2 
	dependency $(v_j^l, v_i^k)$
	if $p = s(v_j^l) = g(v_i^k)$
	and 
	$\hat{t}_g(v_i^k) \leq \hat{t}_g(v_j^l)$. 
	For the base step:
	initially, all \acs{ADG} dependencies 
	have been adhered to
	since $v_i^{1}$ is staged 
	$\forall \; i \in \{1, \dots, N \}$.
	For the inductive step: 
	assuming vertices up until $v_i^{k-1}$ 
	and $v_j^{l-1}$ have 
	been completed in accordance with all \acs{ADG} dependencies,
	it is sufficient to ensure $AGV_i$ and $AGV_j$ 
	will not collide at $\bar{p}$ 
	while completing $v_i^k$ and $v_j^l$ 
	respectively, by ensuring
	$t_s(v_i^k) > t_g(v_j^l)$.
	By line \ref{alg:type_2_dependency_condition}
	of \algref{alg:adg_construction}
	the Type 2 dependency $(v_i^k, v_j^l)$ 
	guarantees 
	$t_s(v_i^k) > t_g(v_j^l)$.
	Since, by \assref{ass:acyclic_ADG}, 
	the \acs{ADG} is acyclic, 
	at least one vertex of the \acs{ADG} 
	can be \textit{in-progress} at all times.
	By the finite nominal execution time
	of the \acs{MAPF} solution in \defref{def:MAPF_solution},
	despite a finite number 
	of delays of finite duration,
	finite-time plan completion is established. 
	This completes the proof.
	\qedproof
\end{myproof}

\section{Switching Dependencies in the \acl{ADG}}
\label{sec:switching_dependencies}

We now introduce the concept of a reversed \acs{ADG} dependency.
In the \acs{ADG}, \textit{Type 2} dependencies essentially encode an ordering constraint 
  for \acs{AGVs} visiting a vertex in $\mathcal{G}$.
The idea is to switch this ordering 
  to minimize the effect an unforeseen delay has on the task completion time of each \acs{AGV}.

\subsection{Reverse \textit{Type 2} Dependencies}
\label{sec:reverse_dependencies}

\begin{figure}
  \centering
  \begin{tikzpicture}
    \begin{scope}[every node/.style={draw=colorborder, fill=colorinside, thick, ellipse, minimum size=2pt}]
        \node (Ai) at (0.0, 1.4) {$v_i^{k-1}$};
        \node (Bi) at (2.4, 1.4) {$v_i^{k}$};
        \node (Ci) at (4.0, 1.4) {$v_i^{k+1}$};
        \node (Aj) at (0.0, 0.0) {$v_j^{l-1}$};
        \node (Bj) at (1.6, 0.0) {$v_j^{l}$};
        \node (Cj) at (4.0, 0.0) {$v_j^{l+1}$};
    \end{scope}
    \begin{scope}[every node/.style={fill=none,thin,draw=none}]
        \node (Li) at (-1.5, 1.4) {};
        \node (Ri) at (5.5, 1.4) {};
        \node (Lj) at (-1.5, 0.0) {};
        \node (Rj) at (5.5, 0.0) {};
        \node (Legend1L) at (-0.5, -1.0) {};
        \node (Legend1R) at (0.5, -1.0) {};
        \node (Legend2L) at (2.0, -1.0) {};
        \node (Legend2R) at (3.0, -1.0) {};
        \node (Legend1) at (1.2, -1.0) {forward};
        \node (Legend2) at (3.7, -1.0) {reverse};
    \end{scope}
    \begin{scope}[>={Stealth[colorborder]},
                  every node/.style={fill=none,circle},
                  every edge/.style={draw=colorborder,very thick}]
        \path [->] (Ai) edge node {} (Bi);
        \path [->] (Bi) edge node {} (Ci);
        \path [->] (Aj) edge node {} (Bj);
        \path [->] (Bj) edge node {} (Cj);
    \end{scope}
    \begin{scope}[>={Stealth[colorborder]},
                  every node/.style={fill=none,circle},
                  every edge/.style={draw=colorborder,very thick, path fading=east}]
        \path [-] (Ci) edge node {} (Ri);
        \path [-] (Cj) edge node {} (Rj);
    \end{scope}
    \begin{scope}[>={Stealth[colorborder]},
                  every edge/.style={draw=colorborder,very thick, path fading=west}]
        \path [->] (Li) edge node {} (Ai);
        \path [->] (Lj) edge node {} (Aj);
    \end{scope}
    \begin{scope}[>={Stealth[red]},
                  every node/.style={fill=none,circle},
                  every edge/.style={draw=red,very thick}]
        \path [->] (Cj) edge node {} (Ai);
        \path [->] (Legend2L) edge node {} (Legend2R);
    \end{scope}
    \begin{scope}[>={Stealth[black]},
                  every node/.style={fill=none,circle},
                  every edge/.style={draw=black,very thick}]
        \path [->] (Bi) edge node {} (Bj);
        \path [->] (Legend1L) edge node {} (Legend1R);
    \end{scope}
  \end{tikzpicture}
  \caption{A subset of an \acs{ADG} with a dependency (black) and its reverse (red)}
  \label{fig:dependency_and_reverse}
\end{figure}
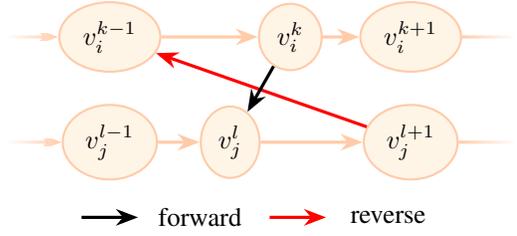

We introduce the notion of a 
  reverse \textit{Type 2} dependency in \defref{def:reverse_dependency}.
It states that a dependency 
  and its reverse encode the same collision avoidance constraints, 
  but with a reversed \acs{AGV} ordering.
\lemref{lem:reversed_dependency} can be used 
  to obtain a dependency which conforms to \defref{def:reverse_dependency}.
\lemref{lem:reversed_dependency} is illustrated graphically in \figref{fig:dependency_and_reverse}.

\begin{mydefinition}[Reverse \textit{Type 2} dependency]
	Consider a \textit{Type 2} dependency $d = (v_i^k, v_{j}^{l})$. $d$ requires $t_\text{s}(v_{j}^{l}) \geq t_g(v_i^k)$.
	A reverse dependency of $d$ is a dependency $d'$ that ensures $t_\text{s}(v_i^k) \geq t_g(v_{j}^{l})$.
	\label{def:reverse_dependency}
\end{mydefinition}

\begin{mylemma}[Reversed \text{Type 2} dependency]
	Let $v^k_i,v^l_j,v^{l+1}_{j},v^{k-1}_{i} \; \in \mathcal{V}_\text{ADG}$. 
	Then $d' = (v^{l+1}_{j}, v^{k-1}_{i})$ is the reverse dependency of $d = (v^k_i,v^l_j)$.
  \label{lem:reversed_dependency}
\end{mylemma}

\begin{myproof}[]
  The dependency $d = (v^k_i,v^l_j)$ 
    encodes the constraint $t_s(v_{j}^{l}) \geq t_g(v_i^k)$.
  The reverse of $d$ 
    is denoted as $d' = (v^{l+1}_{j}, v^{k-1}_{i})$.
  $d'$ encodes 
    the constraint $t_s(v_{i}^{k-1}) \geq t_g(v_j^{l+1})$.
  By definition, 
    $t_s(v_i^k) \geq t_g(v_i^{k-1}) $ 
    and $ t_s(v_j^{l+1}) \geq t_g(v_j^l) $.
  Since $t_g(v) \geq t_s(v)$, 
    this implies that $d'$ 
    encodes the constraint 
    $ t_s(v_i^k) \geq t_g(v_{j}^{l}) $, 
    satisfying \defref{def:reverse_dependency}. \qedproof
\end{myproof}



The modified \acs{ADG} ensures that
  reverse dependencies maintain collision avoidance 
  since adjacent vertices in $\mathcal{V}_\text{ADG}$ refer
  to spatially different locations, cf. line \ref{alg:adg_construction_spatial_exclusivity} in \algref{alg:adg_construction}. 

\subsection{\acl{SADG}}
\label{sec:sadg}

Having introduced reverse \textit{Type 2} dependencies, 
  it is necessary to formalize the manner in which we can 
  select dependencies to obtain a resultant \acs{ADG}.
A cyclic \acs{ADG} implies that two events 
  are mutually dependent, 
  in turn implying a deadlock. 
To ensure deadlock-free plan execution, 
  it is sufficient to ensure that the selected dependencies
  result in an acyclic \acs{ADG}.
Additionally, to maintain the collision-avoidance 
  guarantees implied by the original \acs{ADG},
  it is sufficient to select at least one of the 
  forward or reverse dependencies 
  of each forward-reverse dependency pair 
  in the resultant \acs{ADG}.
Since selecting both a forward and reverse dependency 
  always results in a cycle within the \acs{ADG}, 
  we therefore must either select between the 
  forward or the reverse dependency.
To this end, we formally define a \acf{SADG} in \defref{def:sadg} 
  which can be used to obtain the resultant \acs{ADG} 
  given a selection of forward or reverse dependencies.
  

\begin{mydefinition}[\acl{SADG}]
	Let an \acs{ADG} as in \defref{def:action_dependency_graph} 
	contain $m_T$ forward-reverse dependency 
	pairs determined using \defref{def:reverse_dependency}.
  From this \acs{ADG} we can construct a
    \acl{SADG} 
    $ \textit{SADG}(\textbf{\textit{b}}) : \{0,1\}^{m_T} \to \mathbb{G}$
    where $\mathbb{G}$ is the set of all possible \acs{ADG} graphs
    obtained by the boolean vector 
    $\textbf{\textit{b}} = \{b_1,\dots,b_{m_T}\}$,
	where $b_m = 0$ and $b_m = 1$ imply selecting the forward and reverse 
	dependency of pair $m$ respectively, for $m \in \{1,\dots,m_T\}$. 
	\label{def:sadg}
\end{mydefinition}


\begin{mycorollary}[\acs{SADG} plan execution]
	Consider an \acs{SADG}, $SADG(\textbf{\textit{b}})$, 
	  as in \defref{def:sadg}.
	If \textbf{\textit{b}} is chosen such that 
	  \ADGgraph $= SADG(\textbf{\textit{b}})$ is acyclic, 
	  and no dependencies in \ADGgraph point from vertices 
	  that are \textit{staged} or \textit{in-progress} 
	  to vertices that are \textit{completed}, 
	  \ADGgraph will guarantee collision- 
	  and deadlock-free plan execution.
	\label{col:switchable_ADG_plan_execution}
\end{mycorollary}
\begin{myproof}
	By definition, any \textbf{\textit{b}} 
	  will guarantee collision-free plans, 
	  since at least one dependency of each forward-reverse 
	  dependency pair is selected, by \propref{prop:collision_deadlock_free_plan_exec}.
	If \textbf{\textit{b}} ensures  
	  $\acs{ADG} = \acs{SADG}(\textbf{\textit{b}})$ is acyclic, 
	  and the resultant $\acs{ADG}$ has no dependencies 
	  pointing from vertices 
	  that are \textit{staged} or \textit{in-progress} 
	  to vertices that are \textit{completed}, 
	  the dependencies within the \acs{ADG} are not mutually constraining, 
	  guaranteeing deadlock-free plan execution.
\end{myproof}


The challenge is finding \textbf{\textit{b}} 
  which ensures $\acs{SADG}(\textbf{\textit{b}})$ is acyclic, 
  while simultaneously minimizing the cumulative route completion times
  of the \acs{AGV} fleet. 
This is formulated as 
  an optimization problem in \secref{sec:opt_formulation}. 

\section{Optimization-Based Approach}
\label{sec:opt_formulation}

Having introduced the \acs{SADG}, 
  we now formulate an optimization problem which can be used to
  determine \textbf{\textit{b}} such that the resultant \acs{ADG} 
  is acyclic,
  while minimizing cumulative \acs{AGV} route completion times.  
The result is a \acf{MILP} which we solve in a closed-loop feedback scheme,
  since the optimization problem updates the \acs{AGV} ordering at each iteration 
  based on the delays measured at that time-step.

\subsection{Translating a \acl{SADG} to Temporal Constraints}

\subsubsection{Regular \acs{ADG} Constraints}

Let us introduce the optimization variable $t_{i, s}^k$ 
  which, 
  once a solution to the optimization problem is determined,
  will be equal to $t_s(v_i^k)$. 
The same relation applies to the optimization variable $t_{i, g}^k$ 
  and $t_g(v_i^k)$.
The event-based constraints within the \acs{SADG} 
  can be used in conjunction with a predicted duration of each event to 
  determine when each \acs{AGV} is expected to complete its plan.
Let $\tau(v_i^k)$ be the modeled time it will take $AGV_i$ 
  to complete event $v_i^k \in \mathcal{V}_\text{ADG}$ 
  based solely on dynamical constraints, 
  route distance and assuming the \acs{AGV} is not blocked.
For example, 
  we could let $\tau$ equal the roadmap edge length 
  divided by the expected nominal \acs{AGV} velocity.
We can now specify the temporal constraints 
  corresponding to the \textit{Type 1} dependencies of the plan of $AGV_i$ as
\begin{equation}
  \begin{split}
    t_{i,g}^1 & \geq t_{i,s}^1 + \tau(v_i^1) , \\
    t_{i,s}^2 & \geq t_{i,g}^1 , \\
    t_{i,g}^2 & \geq t_{i,s}^2 + \tau(v_i^2), \\ 
    t_{i,s}^3 & \geq t_{i,g}^2 , \\
    \vdots \hspace{5mm}   &  \hspace{10mm} \vdots \\
    t_{i,s}^{N_i} & \geq t_{i,g}^{N_i-1}  , \\
    t_{i,g}^{N_i} & \geq t_{i,s}^{N_i} + \tau(v_i^{N_i}) . 
  \end{split}
  \label{eq:type_1_temporal_constraints}
\end{equation}

Consider a \textit{Type 2} dependency $(v_i^k, v_{j}^{l})$ 
  within the \acs{ADG}.
This can be represented by the temporal constraint

\begin{equation}
  t_{j,s}^l > t_{i,g}^k,
  \label{eq:type_2_temporal_constraints}
\end{equation}

\noindent where the strict inequality is required to 
  guarantee that $\acs{AGV}_i$ and $\acs{AGV}_j$ 
  never occupy the same spatial region. 

\subsubsection{Adding Switchable Dependency Constraints}
\label{sec:determine_SADG}
We now introduce the temporal constraints 
  which represent the selection of forward or reverse 
  dependencies in the \acs{SADG}.
Initially, consider the 
  set $\mathcal{E}_\text{ADG}^\text{{Type 2}} = \{ e \in \mathcal{E}_\text{ADG} | e \text{ is Type 2} \}$ 
  which represents the sets of all \textit{Type 2} dependencies.
The aim here is to determine 
  a set $\mathcal{E}_\text{ADG}^\text{switchable} \subset \mathcal{E}_\text{ADG}^\text{{Type 2}}$ 
  containing the dependencies 
  which could potentially be switched and form part 
  of the \acs{MILP} decision space. 

\begin{figure}
  \centering
  \includegraphics[width=0.8\linewidth]{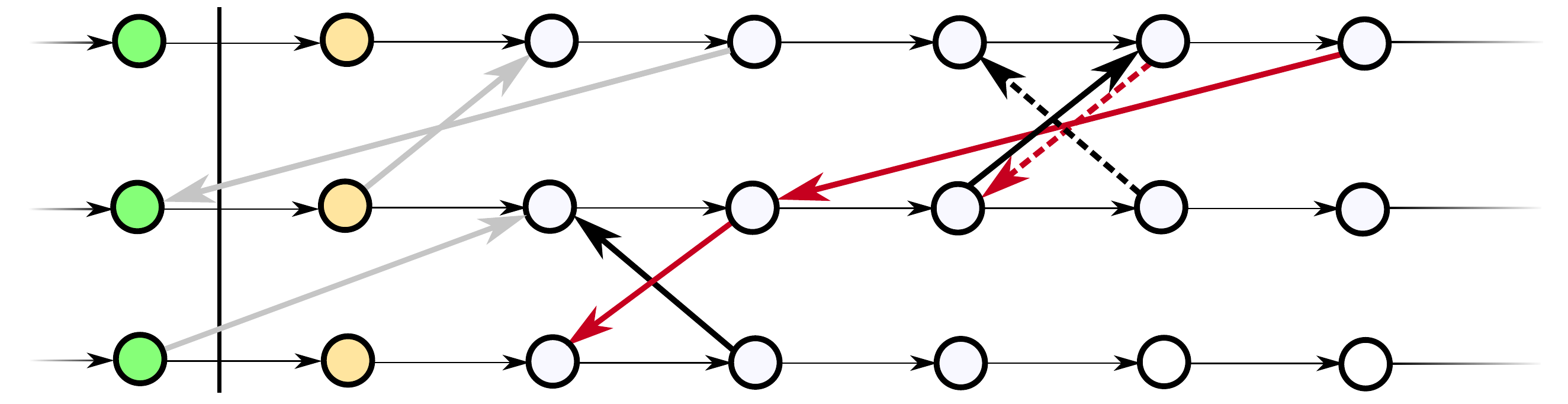}
  \caption{
    Dependencies contained in $\mathcal{E}_\text{ADG}^\text{switchable}$,
    shown in black (forward) and red (reverse). 
    Gray dependencies are in 
    $ \mathcal{E}_\text{ADG}^\text{{Type 2}} \setminus \mathcal{E}_\text{ADG}^\text{switchable} $.
    }
  \label{fig:choosing_dependencies}
\end{figure}

Consider $e_\text{fwd} = (v_f,v_f') \in \mathcal{E}_\text{ADG}^\text{{Type 2}}$
  and its reverse dependency $e_\text{rev} = (v_r,v_r')$.
$e_\text{fwd}$ and $e_\text{rev}$ are 
  contained within $\mathcal{E}_\text{ADG}^\text{switchable}$
  if the status of $v_f,v_f', v_r, v_r'$ is \emph{staged}.
An illustrative example of the dependencies contained within 
  $\mathcal{E}_\text{ADG}^\text{switchable}$ is shown in \figref{fig:choosing_dependencies}.
Having determined $\mathcal{E}_\text{ADG}^\text{switchable}$, 
  the next step is to include the switched dependencies as temporal constraints.
Directly referring to \secref{sec:sadg}, 
  we assume $m_T$ forward-reverse dependency pairs in $\mathcal{E}_\text{ADG}^\text{switchable}$, 
  where the Boolean $b_m$ is used to select 
  the forward or reverse dependency of the
  $m$th forward-reverse dependency pair, 
  $m \in \{ 1,\dots,m_T \}$.
These temporal constraints can be written as

\begin{equation}
  \begin{alignedat}{2}
    t_{j,s}^l & > t_{i,g}^k - b_m M, \; && \\
    t_{i,s}^{k-1} & > t_{j,g}^{l+1} - \big(1 - b_m\big)M, \; &&
  \end{alignedat}
  \label{eq:type_2_switchable_constraints}
\end{equation}
where $M$ is a large, 
  positive constant such that $M > \max_i t_i^{N_i}$. 
Note that $\max_i t_i^{N_i}$ can be approximated 
  by estimating the maximum anticipated delays
  experienced by the \acs{AGVs}.
In practice, however,
  finding such an upper bound on delays is
  not evident, 
  meaning we choose $M$ to be a conservatively high
  value.
  

\subsection{Optimization Problem Formulation}

We have shown that an \acs{SADG} is represented by 
  the temporal constraints in \eqref{eq:type_1_temporal_constraints} 
  through \eqref{eq:type_2_switchable_constraints} 
  for $i \in \{1,\dots,N\}$, $m \in \{1,\dots,m_T\}$.
Minimizing the cumulative route completion time of all \acs{AGVs} 
  is formulated as the following optimization problem

\begin{equation}
	\begin{split}	
		\min_{ \textbf{\textit{b}}, \; \textbf{\textit{t}}_s, \; \textbf{\textit{t}}_g } \hspace{2mm} & \sum_{i=1}^{N} t_{i,g}^{N_i} \\
		\text{s.t. } &  \eqref{eq:type_1_temporal_constraints} \; \forall \; i = \{ 1 , \dots , N \}, \\ %
    & \eqref{eq:type_2_temporal_constraints} \; \forall \; e \in \mathcal{E}_\text{ADG}^\text{{Type 2}} \setminus \mathcal{E}_\text{ADG}^\text{switchable}, \\
    & \eqref{eq:type_2_switchable_constraints} \; \forall \; e \in \mathcal{E}_\text{ADG}^\text{switchable}, \\
	\end{split}
	\label{eq:MILP_objective_function}
\end{equation}
where $\textbf{\textit{b}} : \{0,1\}^{m_T}$ is a vector containing all the binary variables $b_m$ and the vectors $\textbf{\textit{t}}_s$ and $\textbf{\textit{t}}_g$ contain all the variables $t_{i,s}^k$ and $t_{i,g}^k$ respectively $\forall \; k \in \{1,\dots,N_i\}, i \in \{1,\dots,N\}$.

\subsection{Solving the \acs{MILP} in a Feedback Loop}

The aforementioned optimization formulation can be solved  
  based on the current \acs{AGV} positions in a feedback loop.
The result is a continuously updated \ADGgraph 
  which guarantees minimal cumulative route completion times 
  based on current \acs{AGV} delays.
This feedback strategy is defined in \algref{alg:closed-loop-algorithm}.

An important aspect to optimal feedback control strategies 
  is that of recursive feasibility, 
  which means that the optimization problem will 
  remain feasible as long as the control law is applied.
The control strategy outlined in \algref{alg:closed-loop-algorithm} 
  is guaranteed to remain recursively feasible, 
  as formally shown in \propref{prop:recursive_feasibility}.

\begin{algorithm}[]
  \caption{Switching \acs{ADG} Feedback Scheme}
  \begin{algorithmic}[1]
    \renewcommand{\algorithmicrequire}{\textbf{Input:}}
    \renewcommand{\algorithmicensure}{\textbf{Result:}}
    \STATE Get goals and locations
    \STATE Solve \acs{MAPF} to obtain $\mathcal{P}$
    \STATE Construct \acs{ADG} using \algref{alg:adg_construction}
    \STATE Determine $\text{\acs{SADG}}(\textbf{\textit{b}})$ 
      and set $\textbf{\textit{b}}=\textit{\textbf{0}}$ (see \secref{sec:determine_SADG})
    \vspace{1mm}
    \WHILE {Plans not done}
      \STATE get current position along plans for each robot
      \STATE $\textbf{\textit{b}} \leftarrow$ \acs{MILP} in \eqref{eq:MILP_objective_function}
      \STATE $\text{\acs{ADG}} \leftarrow \text{\acs{SADG}}(\textbf{\textit{b}})$
    \ENDWHILE
 \end{algorithmic}
 \label{alg:closed-loop-algorithm}
\end{algorithm}

\begin{myproposition}[Recursive Feasibility]
  Consider an \ac{ADG}, 
    as defined in \defref{def:action_dependency_graph}, 
    which is acyclic at time $t=0$.
  Consecutively applying the \ac{MILP} solution  
    from \eqref{eq:MILP_objective_function} 
    is guaranteed to ensure 
    the resultant \ac{ADG} remains 
    acyclic for all $t > 0$.
  \label{prop:recursive_feasibility}
\end{myproposition}


\begin{myproof}[]
  Proof by induction. 
  Consider an acyclic \acs{ADG} 
	as defined in \defref{def:action_dependency_graph},
	at a time $t$. 
  The \acs{MILP} in \eqref{eq:MILP_objective_function} 
    always has the feasible solution $\textbf{\textit{b}}=0$ 
    if the initial \acs{ADG} 
    (from which the \acs{MILP}'s constraints in \eqref{eq:type_1_temporal_constraints} through \eqref{eq:type_2_switchable_constraints} are defined) 
    is acyclic.
  Any improved solution of the \acs{MILP} with $\textbf{\textit{b}} \neq 0$ is necessarily feasible, 
    implying a resultant acyclic \acs{ADG}.
  This implies that the \acs{MILP} is guaranteed to 
    return a feasible solution, the resultant \acs{ADG} 
    will always be acyclic if the \acs{ADG} 
    before the \acs{MILP} was solved, 
    was acyclic.
  Since the \acs{ADG} at $t = 0$ is acyclic 
    (a direct result of a \acs{MAPF} solution), 
    it will remain acyclic for $t > 0$. \qedproof
\end{myproof}


\subsection{Decreasing Computational Effort}
\label{sec:decrease_comp_effort}

The time required to solve the \acs{MILP} will directly 
  affect the real-time applicability of this approach.
In general,
  the complexity of the \acs{MILP} increases exponentially in the number of binary variables.
To render the \acs{MILP} less computationally demanding, 
  it is therefore most effective to decrease the number of binary variables.
We present two complementary methods to achieve this goal.

\subsubsection{Switching Dependencies in a Receding Horizon}
\label{sec:reducing_decision_space}

\begin{figure}[]
	\centering
	\includegraphics[width=0.9\linewidth]{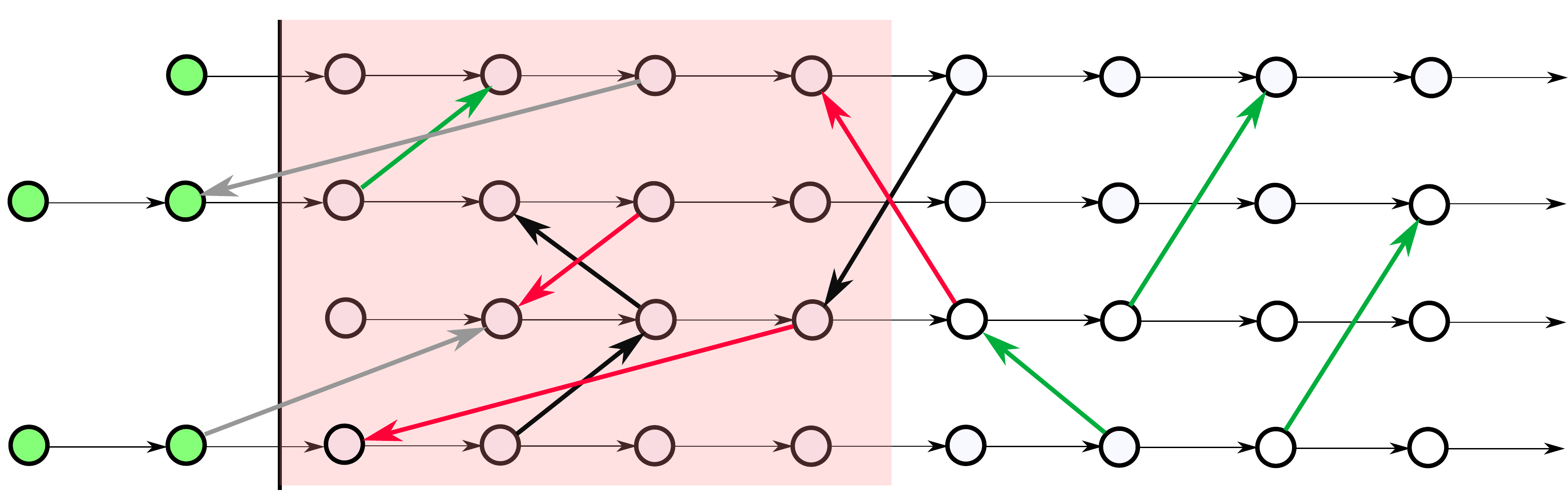}
	\caption{Dependency selection for a horizon of $4$ vertices. 
		Switchable dependency pairs are shown in black (forward) and red (reverse). 
		Regular dependencies considered in the \acs{MILP} are green.
		Dependencies not considered are gray.}
	\label{fig:receding_horizon_dependencies}
\end{figure}

Instead of including all switchable dependency pairs in the 
  set $\mathcal{E}_\text{ADG}^\text{switchable}$,
we can only include the switchable dependencies associated with
  vertices within a horizon $H$ from the last \emph{completed} vertex.  
An illustration of such selection for $H = 4$ 
  is shown in \figref{fig:receding_horizon_dependencies}.
Note that dependency selection using this approach maintains 
  \acs{ADG} acyclicity as in the infinite horizon case,
  because the set $\mathcal{E}_\text{ADG}^\text{switchable}$ 
  is smaller, but \eqref{eq:MILP_objective_function}
  remains recursively feasible since the trivial solution 
  guarantees a acyclic \acs{ADG} at every time-step.
\propref{prop:recursive_feasibility} is equally valid when 
  only considering switchable dependencies in a receding horizon.
The horizon length $H$ can be seen as a tuning parameter which 
  can offer a trade-off between computational complexity and 
  solution optimality.

Note that, to guarantee recursive feasibility,
  any switchable dependencies which are not within the horizon $H$
  (e.g. the green dependencies in \figref{fig:receding_horizon_dependencies})
  still 
  need to be considered within the \acs{MILP}
  by applying the constraint 
  in \eqref{eq:type_2_temporal_constraints}.
Future work will look into a receding horizon approach that does not 
  necessarily require the consideration of all these constraints
  while guaranteeing recursive feasibility.

\subsubsection{Dependency Grouping}
\label{sec:dependency_grouping}

\begin{figure}
  \centering
  \begin{subfigure}{0.23\textwidth}
    \centering
    \begin{tikzpicture}
      \begin{scope}[every node/.style={draw=colorborder, fill=colorinside, thick, ellipse, minimum size=2pt}]
        \node (Ai) at (0.0, 1.0) {};
        \node (Bi) at (1.0, 1.0) {};
        \node (Ci) at (2.0, 1.0) {};
        \node (Di) at (3.0, 1.0) {};
        \node (Aj) at (0.0, 0.0) {};
        \node (Bj) at (1.0, 0.0) {};
        \node (Cj) at (2.0, 0.0) {};
        \node (Dj) at (3.0, 0.0) {};
      \end{scope}
      \begin{scope}[every node/.style={fill=none, thin, draw=none}]
        \node (Li) at (-0.5, 1.0) {};
        \node (Ri) at (3.5, 1.0) {};
        \node (Lj) at (-0.5, 0.0) {};
        \node (Rj) at (3.5, 0.0) {};
      \end{scope}
      \begin{scope}[>={Stealth[colorborder]},
                    every node/.style={fill=none,circle},
                    every edge/.style={draw=colorborder,very thick}]
          \path [->] (Ai) edge node {} (Bi);
          \path [->] (Bi) edge node {} (Ci);
          \path [->] (Ci) edge node {} (Di);
          \path [->] (Aj) edge node {} (Bj);
          \path [->] (Bj) edge node {} (Cj);
          \path [->] (Cj) edge node {} (Dj);
      \end{scope}
      \begin{scope}[>={Stealth[colorborder]},
                    every node/.style={fill=none,circle},
                    every edge/.style={draw=colorborder,very thick, path fading=east}]
          \path [-] (Di) edge node {} (Ri);
          \path [-] (Dj) edge node {} (Rj);
      \end{scope}
      \begin{scope}[>={Stealth[colorborder]},
                    every edge/.style={draw=colorborder,very thick, path fading=west}]
          \path [->] (Li) edge node {} (Ai);
          \path [->] (Lj) edge node {} (Aj);
      \end{scope}
      \begin{scope}[>={Stealth[red]},
                    every node/.style={fill=none,circle},
                    every edge/.style={draw=red, thick}]
          \path [->] (Ci) edge node {} (Aj);
          \path [->, dashed] (Di) edge node {} (Bj);
      \end{scope}
      \begin{scope}[>={Stealth[black]},
                    every node/.style={fill=none,circle},
                    every edge/.style={draw=black, thick}]
          \path [->] (Bj) edge node {} (Bi);
          \path [->, dashed] (Cj) edge node {} (Ci);
      \end{scope}
    \end{tikzpicture}
    \caption{\textit{same}}
    \label{fig:same_DG}
  \end{subfigure}
  \begin{subfigure}{0.23\textwidth}
    \centering
    \begin{tikzpicture}
      \begin{scope}[every node/.style={draw=colorborder, fill=colorinside, thick, ellipse, minimum size=2pt}]
        \node (Ai) at (0.0, 1.0) {};
        \node (Bi) at (1.0, 1.0) {};
        \node (Ci) at (2.0, 1.0) {};
        \node (Di) at (3.0, 1.0) {};
        \node (Aj) at (0.0, 0.0) {};
        \node (Bj) at (1.0, 0.0) {};
        \node (Cj) at (2.0, 0.0) {};
        \node (Dj) at (3.0, 0.0) {};
      \end{scope}
      \begin{scope}[every node/.style={fill=none, thin, draw=none}]
        \node (Li) at (-0.5, 1.0) {};
        \node (Ri) at (3.5, 1.0) {};
        \node (Lj) at (-0.5, 0.0) {};
        \node (Rj) at (3.5, 0.0) {};
      \end{scope}
      \begin{scope}[>={Stealth[colorborder]},
                    every node/.style={fill=none,circle},
                    every edge/.style={draw=colorborder,very thick}]
          \path [->] (Ai) edge node {} (Bi);
          \path [->] (Bi) edge node {} (Ci);
          \path [->] (Ci) edge node {} (Di);
          \path [->] (Aj) edge node {} (Bj);
          \path [->] (Bj) edge node {} (Cj);
          \path [->] (Cj) edge node {} (Dj);
      \end{scope}
      \begin{scope}[>={Stealth[colorborder]},
                    every node/.style={fill=none,circle},
                    every edge/.style={draw=colorborder,very thick, path fading=east}]
          \path [-] (Di) edge node {} (Ri);
          \path [-] (Dj) edge node {} (Rj);
      \end{scope}
      \begin{scope}[>={Stealth[colorborder]},
                    every edge/.style={draw=colorborder,very thick, path fading=west}]
          \path [->] (Li) edge node {} (Ai);
          \path [->] (Lj) edge node {} (Aj);
      \end{scope}
      \begin{scope}[>={Stealth[red]},
                    every node/.style={fill=none,circle},
                    every edge/.style={draw=red, thick}]
          \path [->] (Di) edge node {} (Aj);
          \path [->, dashed] (Ci) edge node {} (Bj);
      \end{scope}
      \begin{scope}[>={Stealth[black]},
                    every node/.style={fill=none,circle},
                    every edge/.style={draw=black, thick}]
          \path [->, dashed] (Cj) edge node {} (Bi);
          \path [->, shift left=.4ex] (Bj) edge node {} (Ci);
      \end{scope}
    \end{tikzpicture}
    \caption{\textit{opposite}}
    \label{fig:opposite_DG}
  \end{subfigure}
  \caption{Dependency groups. Each dependency is either original (black) or reversed (red). 
  Reverse and forward dependency pairings are differentiated by line styles.}
  \label{fig:dependency_groups}
\end{figure}
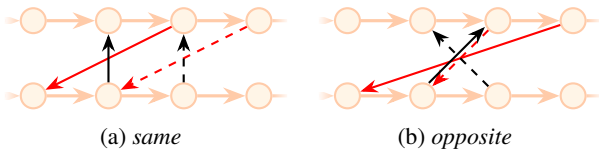

We observed that multiple dependencies would often form patterns, 
  two of which are shown in \figref{fig:dependency_groups}.
These patterns are referred to as \textit{same-direction} 
  and \textit{opposite-direction} dependency groups, 
  shown in \figref{fig:same_DG} and \figref{fig:opposite_DG} respectively.
These groups share the same property that the resultant \acs{ADG} is acyclic 
  if and only if either all the forward or all the reverse dependencies are active.
This means that a single binary variable is 
  sufficient to describe the switching of all the dependencies within the group, 
  decreasing the variable space of the \acs{MILP} in \eqref{eq:MILP_objective_function}. 
Once such a dependency group has been identified, 
  the temporal constraints can then be defined as

\begin{equation}
	\begin{alignedat}{2}
		t_{j,s}^l & > t_{i,g}^k - b_{DG}M \; && \forall \; (v_i^k, v_j^l) \in \mathcal{DG}_\text{fwd}, \\
		t_{j,s}^l & > t_{i,g}^k - (1 - b_{DG})M \; \; && \forall \; (v_i^k, v_j^l) \in \mathcal{DG}_\text{rev},
	\end{alignedat}
\end{equation}

\noindent where $\mathcal{DG}_\text{fwd}$ and $\mathcal{DG}_\text{rev}$ refer to 
  the forward and reverse dependencies of a particular grouping respectively, 
  and $b_{DG}$ is a binary variable which switches all the forward or reverse 
  dependencies in the entire group simultaneously.



\section{Evaluation}
\label{sec:evaluation}

We design a set of simulations to evaluate the approach in terms of re-ordering efficiency  
  when \acs{AGVs} are subjected to delays while following their initially planned paths.
We use the method presented by H\"onig \etal \cite{hoenigPersistentRobustExecution2018} 
  as a comparison baseline.
All simulations were conducted on a Lenovo Thinkstation 
  with an Intel\textregistered{} Xeon E5-1620 \@ 3.5GHz processor and 64 GB of RAM.

\subsection{Simulation Setup}

The simulations consider a roadmap as shown in \figref{fig:agvs_on_map}.
A team of \acs{AGVs} of size $\{ 30,40,50,60,70 \}$ are each initialized 
  with a random start and goal position.
\acs{ECBS} \cite{barerSuboptimalVariantsConflictBased2014} is used 
  to solve the \acs{MAPF} with sub-optimality factor $w = 1.6$.
We consider delays of duration $k = \{1,3,5,10,15,20,25\}$ time-steps.
At the $k^{th}$ time-step, 
  a random subset (20\%) of the \acs{AGVs} are stopped for a length of $k$.
\eqref{eq:MILP_objective_function} is solved at each time-step with $M=10^4$.
We evaluate our approach using a Monte Carlo method: 
  for each \acs{AGV} team size  
  and delay duration configuration, 
  we consider $100$ different randomly selected goal/start positions.
The receding horizon dependency selection 
  and dependency groups are used 
  as described in \secref{sec:decrease_comp_effort}.

\subsection{Performance Metric and Comparison}

Performance is measured by considering 
  the cumulative plan completion time of all the \acs{AGVs}.
This is compared to the same metric 
  using the original \acs{ADG} approach 
  with no switching as in \cite{hoenigPersistentRobustExecution2018},
  which is equivalent to forcing 
  the solution of \eqref{eq:MILP_objective_function} 
  to $\textbf{\textit{b}} = \textbf{0}$ at every time-step.
The \textit{improvement} is defined as

$$ \text{improvement} = \frac{\sum t_\text{baseline} - \sum t_\text{switching}}{\sum t_\text{baseline}} \cdot 100 \%, $$

\noindent where $\sum t_\ast$ refers to the cumulative plan completion time for all \acs{AGVs}.
Note that we consider cumulative plan completion time instead of the make-span 
  because we want to ensure each \acs{AGV} completes its plans as soon as possible, 
  such that it can be assigned a new task.

Another important consideration is the time it takes to solve the \acs{MILP} in \eqref{eq:MILP_objective_function} at each time-step.
For our simulations, 
  the \acs{MILP} was solved using the academically orientated 
  \ac{CBC} solver \cite{john_forrest_2018_1317566}.
However, 
  based on preliminary tests, 
  we did note better performance 
  using the commercial solver Gurobi \cite{gurobi}. 
This yielded computational time 
  improvements by 
  a factor 1.1 up to 20.

\subsection{Results and Discussion}

\begin{figure}[]
  \centering
  \includegraphics[width=\linewidth]{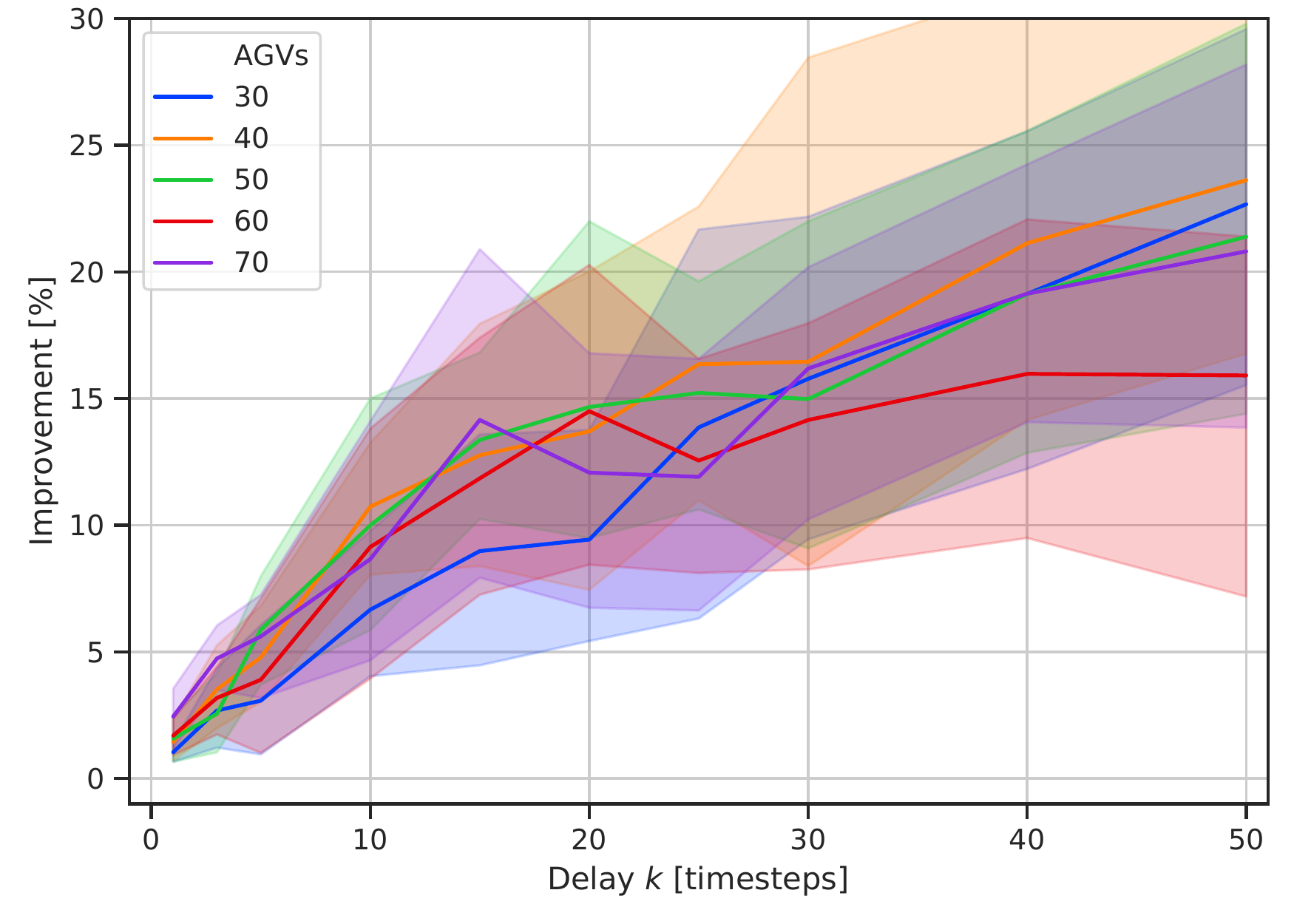}
  \caption{
    Average improvement of $100$ scenarios for various delay lengths and \acs{AGV} group sizes. 
    Each scenario refers to different randomly generated starts/goals 
      and a randomly selected subset of delayed \acs{AGVs}. 
    Solid lines depict the average, 
      lighter regions encapsulate the min-max values.
  }
  \label{fig:intial_results}
\end{figure}

To showcase the efficacy of our approach, 
  we first determine the average improvement of $100$ random scenarios 
  using the minimum switching horizon length of $1$ 
  for different \acs{AGV} team sizes and delay lengths, 
  shown in \figref{fig:intial_results}. 
The average improvement is highly correlated to the delay duration 
  experienced by the \acs{AGVs}.
  
\begin{figure}[]
	\centering
	\begin{subfigure}{\linewidth}
		\includegraphics[width=\linewidth]{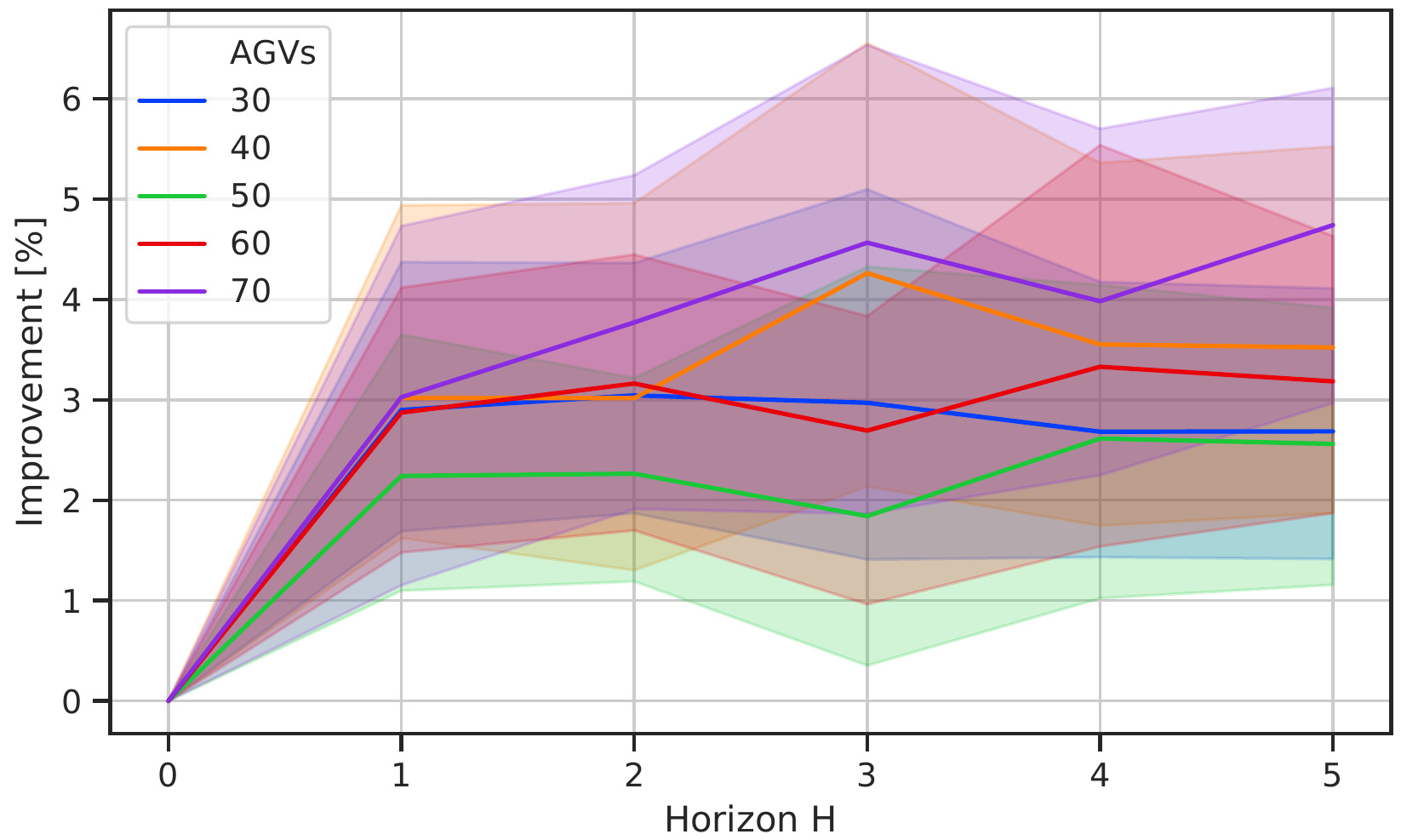}
		\caption{
			Delay $k = 3$.
		}
		\label{fig:horizon_to_performance_k_1}
	\end{subfigure}
	\begin{subfigure}{\linewidth}
		\includegraphics[width=\linewidth]{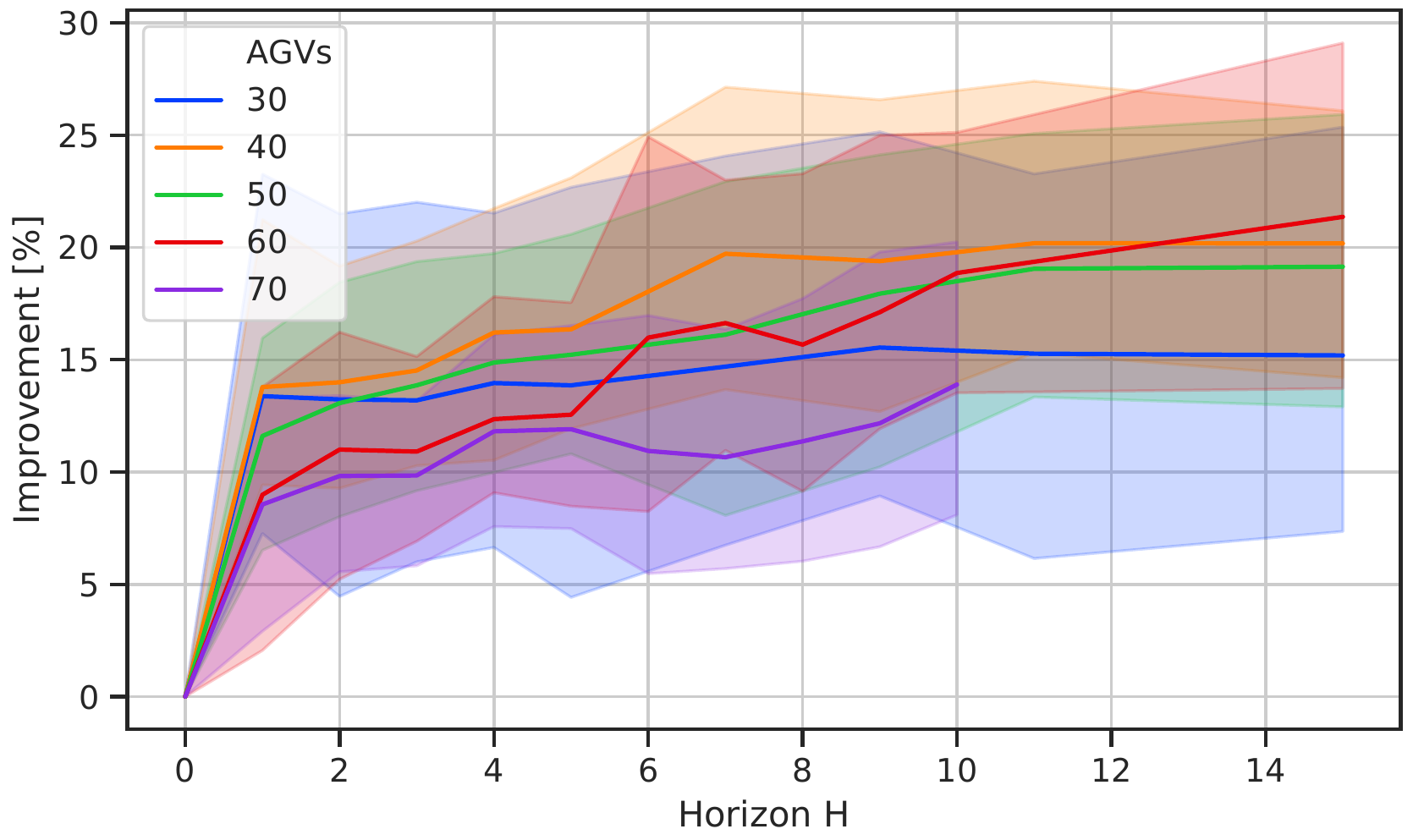}
		\caption{
			Delay $k = 25$.
		}
		\label{fig:horizon_to_performance_k_10}
	\end{subfigure}
	\caption{
		Average improvement of $100$ random start/goal positions
		and delayed \acs{AGV} subset, 
		for different 
		switching horizon lengths,  
	    for different \acs{AGV} group sizes.
	  Solid lines depict the average, 
	    lighter regions encapsulate the min-max values.}
	\label{fig:horizon_to_performance}
\end{figure}

Considering \figref{fig:intial_results}, it is worth noting how the graph layout and \acs{AGV}-to-roadmap density affects the results: 
  the \acs{AGV} group size of $40$ shows 
  the best average improvement for a given delay duration.
This leads the authors to believe there is an 
  optimal \acs{AGV} group size for a given roadmap, 
  which ensures the workspace is both:
\begin{enumerate}
  \item Not too congested to make switching of dependencies impossible 
    due to the high density of \acs{AGVs} occupying the map.
  \item Not too sparse such that switching is never needed since \acs{AGVs} are distant from each other, 
    meaning that switching rarely improves task completion time.
\end{enumerate}

Considering the switching dependency horizon, 
  \figref{fig:horizon_to_performance} shows the 
  average improvement for $100$ random start/goal positions 
  and delayed \acs{AGV} subset selection. 
We observe that a horizon length of $1$ already significantly
  improves performance, and larger horizons seem to gradually 
  increase performance for larger \acs{AGV} teams.

\begin{figure}[]
  \centering
  \includegraphics[width=\linewidth]{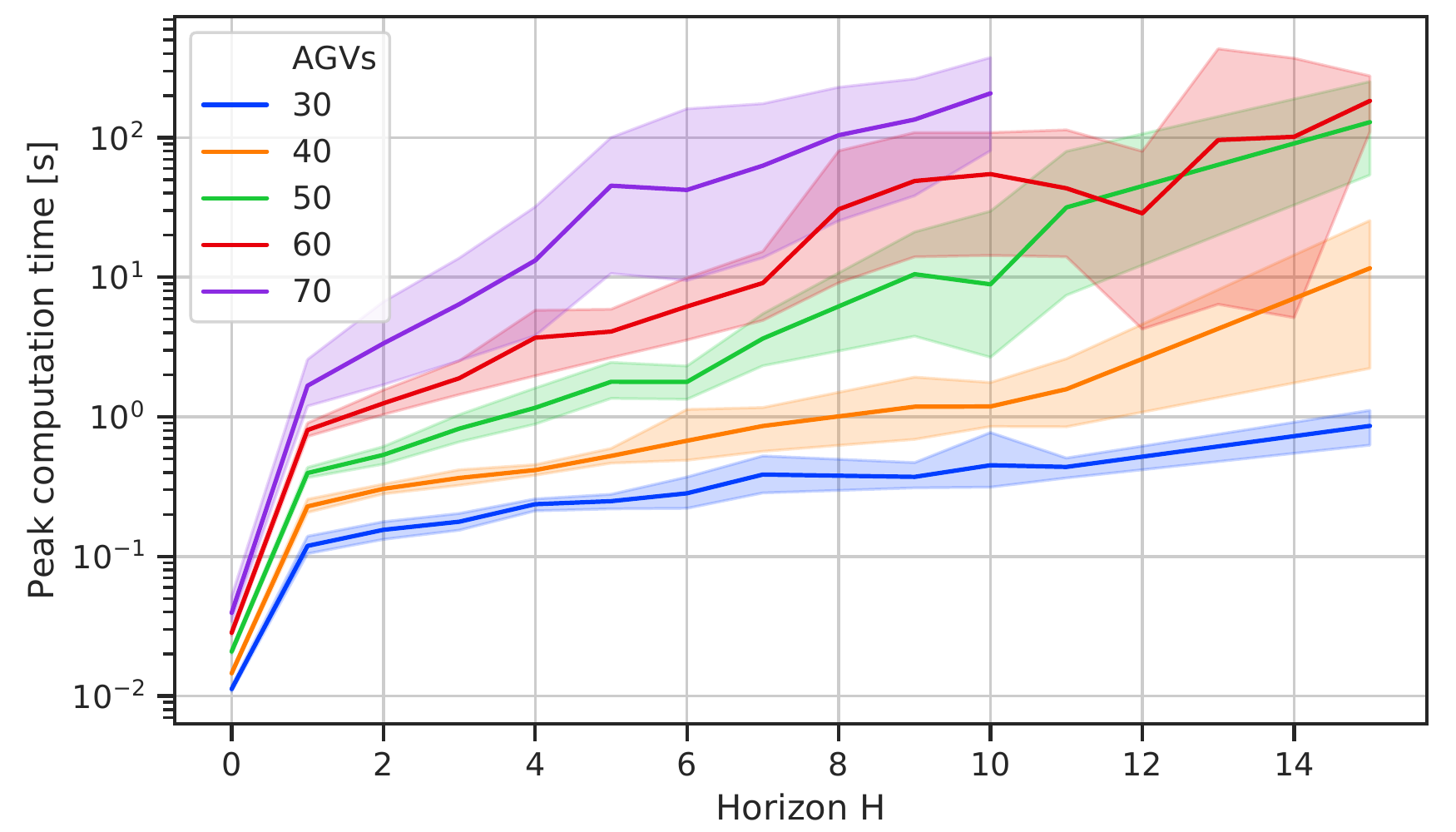}
  \caption{
    The peak computation to solve the optimization problem 
    for different \acs{AGV} team sizes and considered dependency horizon lengths. 
    This plot considers the average improvement for delays $k = \{1,3,5,10,15,20,25\}$.
    Solid lines depict the average, 
        lighter regions encapsulate the min-max values.
  }
  \label{fig:computation_time}
\end{figure}

\figref{fig:computation_time} shows the peak computation time 
  for various horizon lengths and \acs{AGV} team sizes. 
As expected, 
  the computation time is exponential with horizon size 
  and \acs{AGV} team size.
Two additional observations that were made:
\begin{enumerate}
  \item \textit{High variability in results.}
  Note the high variability in improvement indicated by the large lighter regions in \figref{fig:intial_results}.
  This means that for different random start/goal and delay configurations, 
    the improvement varied significantly.
  This is due to the fact that each start/goal combination 
    provides differing degrees-of-freedom 
    from an \acs{ADG} switching perspective.
  \item \textit{Occasional worse performance.}
  Occasionally, albeit rarely, 
    our approach would yield a negative improvement 
    for a particular random start/goal configuration.
  This was typically observed for small delay durations.
  The reason is that the optimization problem solves the 
    switching assuming no future delays.
  However, it may so happen that the \acs{AGV} 
    which was allowed ahead of another, 
    is delayed in the near future, 
    additionally delaying the \acs{AGV} it surpassed.
  We believe a robust optimization approach 
    could potentially resolve this.
\end{enumerate}

\noindent Finally, we emphasize that our proposed approach:
\begin{enumerate}
  \item Is a complementary approach which could be used together with 
          the methods presented in 
          \cite{atzmonRobustMultiAgentPath2020,hoenigPersistentRobustExecution2018}
          and other works.
  \item Applies to directional and weighted roadmaps (since \ac{ADG} switching retains 
          the direction of the original \ac{MAPF} plan);
  \item Persistent planning schemes as in \cite{hoenigPersistentRobustExecution2018},
    as long as all \acs{AGV} plans $\mathcal{P}$ are known when the \acs{ADG} is constructed.
\end{enumerate}


\section{Conclusions and Future Work}
\label{sec:conclusion}

In this paper, we introduced a novel method which,
  given a \acs{MAPF} solution, 
  can be used to switch the ordering of \acs{AGVs} 
  in an online fashion 
  based on currently measured \acs{AGV} delays.
This switching was formulated as 
  an optimization problem as part of a feedback control scheme, 
  while maintaining the deadlock- and collision-free 
  guarantees of the original \acs{MAPF} plan.
Results show that our approach clearly 
  improves the cumulative task completion time of the \acs{AGVs}
  when a subset of \acs{AGVs} are subjected to delays.

In future work, 
  we plan to consider a receding horizon optimization approach.
In this work, \acs{ADG} dependencies 
  can be switched in a receding horizon fashion, 
  but the plans still need to be of finite length 
  for the optimization problem to be formulated.
For truly persistent plans (theoretically infinite length plans), 
  it is necessary to come up with a receding horizon optimization formulation 
  to apply the method proposed in this paper.

Another possible extension is complementing 
  our approach with a local re-planning method.
This is because our approach maintains 
  the originally planned trajectories of the \acs{AGVs}.
However, we observed that under large delays, 
  the originally planned routes can become largely inefficient
  due to the fact the \acs{AGVs} are in entirely different locations along their planned path. 
This could potentially be addressed by introducing local re-planning of trajectories. 

To avoid the occasional worse performance, 
  we suggest a robust optimization approach  
  to avoid switching dependencies which could 
  have a negative impact on the plan execution
  given expected future delays.
  
Finally, to further validate this approach, 
  it is desirable to move towards system-level tests 
  on a real-world intralogistics setup.


%


\bibliographystyle{aaai} 
\bibliography{main}

\end{document}